# Analyze the Robustness of Classifiers under Label Noise

Author: Cheng Zeng czen8507@uni.sydney.edu.au, Yixuan Xu yixu9725@uni.sydney.edu.au, Jiaqi Tian jtia3555@uni.sydney.edu.au

## Abstract

The purpose of this article is to conduct research on the robustness of label noise classifiers to improve the model's robustness to noisy data and ensure its robust- ness in complex actual scenarios. Label noise refers to the situation where labels appear wrong or inaccurate in supervised learning [1]. Recent developments indicate that the impact of label noise on model performance has become a focus of research, especially in practical applications. Therefore, the common problem of inaccurate labels in training data poses a significant challenge. We adopt adversarial machine learning (AML) and importance reweighting methods to deal with the label noise problem and use convolutional neural networks (CNN) as the baseline model. We adjusted the parameters of each sample in each training data to make the model pay more attention to samples that have a significant impact on performance. Such a combination aims to improve the model's adaptability to label noise.

On the CIFAR, FashionMnist0.5, and FashionMnist0.6 datasets, we use evaluation metrics such as precision, recall, and F1 score. By modifying the parameters and conducting multiple experiments, the experimental results show that the two methods of AML and importance reweighting indeed significantly improve the model's resistance to label noise and improve the classification accuracy and robustness.

In summary, our study highlights the importance of addressing label noise in supervised learning. Both methods, adversarial machine learning and importance reweighting, have been shown to be effective in improving the robustness of label noise classifiers. In the future, these insights may provide ideas for improving the reliability and generalization ability of supervised learning models in real-world scenarios.

## 1, Introduction

The robustness of machine learning to noisy data is indispensable in practical applications. The purpose of our paper is to design and implement a classifier that is robust to noise, which can significantly degrade the performance of predictive models. In the field of machine learning, label noise can come from a variety of sources, including human error in the data labeling process, ambiguity in the data itself, or problems in data collection and processing. Furthermore, the real-world impact is huge and good image labeling often requires expensive labor [1]. Due to limited budgets, machine learning models can be misled by noisy data sets and remember wrong relationships [1]. In the field of label noise research, a variety of classifiers and methods exist to improve model performance in the presence of label errors. A common approach is to improve the quality of the data set through noise filtering or correction. This may involve using some outlier detection techniques to exclude samples from the training set that may contain incorrect labels. Although existing research has pro- posed several methods to deal with label noise, they usually have limitations. Examples include the need for a clean validation set or the inability to handle high noise levels in real-world environments, as labels or annotations are often noisy in real-world environments and imperfect real-world scenarios.

We wish to build robust learning algorithms with theoretical guarantees to handle noisy labels. And can handle various levels of label noise without requiring the original validation set. In terms of image noise robust classifiers, some methods also focus on handling specific types of image noise, such as occlusions, blurs, or distortions. These methods may involve augmentation or adversarial training of images to make the model more robust to noise. In this paper, we try to use adversarial machine learning (AML) and importance reweighting methods to solve the label noise problem and choose convolutional neural network (CNN) as the baseline model. We will provide a structured overview and approach, followed by an empirical evaluation of its performance based on four metrics: precision, recall, accuracy, and F1 score.

However, current research still faces some challenges. Real-world labeling errors are often complex and diverse and difficult to capture with simple models. At the same time, some previous works may be based on overly idealized assumptions about noise, limiting their applicability in real scenarios. The purpose of this report is to provide insight into the current status of this problem and to provide substantive data for future research through reproduction and experimentation. According to the four indicators of precision, recall rate, accuracy rate, and F1 score, the results will have different performance when the parameters are modified. We hope to gain a deeper understanding of image noise robustness and provide more reliable solutions for practical applications.

## 2, Previous Work

The impact of label noise on model robustness is an important research topic. Convolutional neural network (CNN) is a basic model in the research. The two models we adopt this time are adversarial machine learning (AML) and importance reweighting methods. Adversarial machine learning (AML) aims to improve a model's ability to resist adversarial attacks. The intuition is that by introducing carefully designed adversarial examples into the training data, the model is made more robust and able to better generalize to unseen samples. The method's generative adversarial networks (GANs) can be used to generate adversarial examples to improve the classifier's tolerance to noise and interference.[2] Assuming that the generator network is G, the discriminator network is D, the input sample is x, and the adversarial example is $x_{adv}$, the process of generating adversarial examples can be expressed by the following formula:

$$X_{adv} = x + \epsilon \cdot sign(\nabla_x J(G(x)), y_{true}) \tag{1}$$

Importance reweighting methods aim to adjust the weight of data samples so that more attention is paid to samples that have a greater impact on model performance during training. This is particularly useful when dealing with issues such as label noise. These methods redistribute weights based on the importance of samples, typically by considering the model's prediction error or uncertainty for each sample. This allows the model to pay more attention to samples that are difficult to classify or are susceptible to noise during the training process [3]. The importance weight can be defined by the relationship between the predicted probability of the sample and the true label:

$$w_i = \frac{1}{|y_{true} - y_{pred}| + \epsilon} \tag{2}$$

In recent years, previous research has made significant progress in solving the problem of robustness to label noise. A series of new algorithms are proposed to deal with the label noise problem. These algorithms not only detect label errors more effectively, but also provide more accurate correction methods. For example, semi-supervised learning methods and active learning methods based on graphical models have become key technologies in the field of label noise robustness. At the same time, the popularity of hardware such as GPUs and TPUs and the optimization of deep learning frameworks allow us to more quickly experiment and iterate new algorithms to increase the robustness of label noise. Label noise robustness is not only applicable to the field of computer vision, but also involves natural language processing, bioinformatics, and other fields [2]. The emergence of these new application areas demonstrates the importance and broad applicability of label noise robustness research.

We try to use adversarial machine learning (AML) and importance reweighting method, CIFAR, FashionMNIST0.5 and FashionMNIST0.6 datasets to improve the robustness of the model in actual scenarios. Through the study of related work on label noise robustness, the status and challenges in this field can be better understood.

## 3, Methods

### 3.1, Noise Rate Estimation Methods

As we mentioned, in practical applications, the data will contain some label noise for many reasons. To solve this problem, we first need to understand the difference between the distribution of noise labels and the distribution of true labels. We usually call the probability that a label is mislabeled the flip rate. The formula is

$$\rho_Y(X) = P(\hat{Y}|Y, X) \tag{3}$$

where X is feature, Y is true label, $\hat{Y}$ is noisy label [3].

In a binary classification problem, we use the following two formulas to represent the probability that the true label is 1 but the noisy label is -1 and the probability that the true label is -1 but the noisy label is 1.

$$\rho_{+1}(X) = P(\hat{Y} = -1|Y = 1, X) \tag{4}$$

$$\rho_{-1}(X) = P(\hat{Y} = 1|Y = -1, X) \tag{5}$$

For multiple classification, we need know all flip rate $f_{ij}$, where $f_{ij}$ represents the probability of true label is i but is incorrectly labeled j. And Transition Matrix:

$$\begin{pmatrix} f11 & f12 & \cdots & f1j \\ f21 & f22 & \cdots & f2j \\ \vdots & \vdots & \ddots & \vdots \\ fi1 & fi2 & \cdots & fij \end{pmatrix}$$

We can use the Transition matrix to adjust the algorithm so that the model can be more robust. According to research[11], transition matrix can be used to adjust the loss function to improve the model's robustness on noisy label data.

In real life, we sometimes do not know the Transition Matrix of data set. So, we need an algorithm to evaluate the transition matrix of the data. According to research [12], the transition matrix can be estimated by analyzing the relationship between the labels predicted by the pre-trained model and the true labels. Specific steps are as follows: 1. We first train a simple nn model on the training dataset. 2. Use the model to predict all data and obtain the probability that each sample belongs to each label. 3. Collect true labels in the test dataset. 4. For each label, find all probabilities predicted by the model to be other labels. 5. Repeat steps 2-4 to calculate the average value of probability to obtain the transition matrix.

Then, I will give an example to help understand this method.

Suppose there are 3 noisy labels: 0, 1, and 2 respectively. First, we find all samples with noise label 0 from the data set. Then we use the model to predict these samples, and the result may be [0.7, 0.2, 0.1] for the label [0, 1, 2]. In this way, we can get the probability that each noise label is predicted to be other labels. The average of these probabilities is then calculated. This average reflects the flip rate. By repeating this process for all noise labels, we can get a complete Transition Matrix.

The formula for estimate transition matrix:

$$T_{ij} = \frac{1}{|S_j|} \sum_{x \in S_j} P(y = i|x) P \tag{6}$$

Where, $S_j$ is the set of samples with all observed noise labeled j, $|S_j|$ is number of sets, $T_{ij}$ is the element in row i and column j of the transition matrix. $T_{ij}$ represents the average probability that the noise label j is predicted by the model to be true label i.

The following pseudo code shows a function for evaluating a Transition matrix:

---
**Algorithm 1** Estimate Transition Matrix with Forward Learning
---
**Require:** Pretrained model $M$, DataLoader $L$
    Set model to evaluation mode $M$.eval()
    Initialize empty list predicted_probabilities ← [ ]
    Initialize empty list observed_labels ← [ ]
    **for** each batch $(\mathbf{x}, \mathbf{y})$ in $L$ **do**
        Note: Forward learning without gradient computation
        **outputs** ← $M(\mathbf{x})$
        **probabilities** ← softmax(**outputs**)
        Add **probabilities** to predicted_probabilities
        Add $\mathbf{y}$ to true_labels
    **end for**
    $\mathbf{P}$ ← concatenate all elements in predicted_probabilities
    $\mathbf{S}$ ← concatenate all elements in true_labels
    Initialize transition matrix $\mathbf{T}$ ← $\mathbf{0}$ with dimensions $(C, C)$ where $C$ is the number of classes
    **for** each class label $j$ **do**
        mask ← $\mathbf{S} == j$
        **for** each class label $i$ **do**
            Average probability of true label $i$ given noisy label $j$
            $T_{ij}$ ← mean($\mathbf{P}_{\mathbf{mask},i}$)
        **end for**
    **end for**
    **return** $\mathbf{T}^\top$
---

We will evaluate the difference between the predicted transition matrix and the real transition matrix in 4.3 section.

In our experiment, we have 3 data sets. The first two data sets (FashionMNIST0.5 and FashionMNIST0.6) already provide a Transition Matrix. The CIFAR data set does not provide a Transition Matrix. So when processing the CIFAR data set, we need to first calculate the Transition Matrix of the data set. When processing other **label noise data for which the flip rate is not known**, you can also use this function to estimate the Transition Matrix of the data set before other steps.

### 3.2, Method1: CNN with importance reweighting

In order to design a classifier that is robust to label noise, we should consider many factors, such as preprocessing, model selection, and noise processing. According to research [4], normalization can make the loss function robust to noisy labels. Therefore, choosing normalization as a data preprocessing method can improve the performance of the model under noisy data. According to research [5], CNN can effectively learn features of images through convolutional layers, pooling layers, and fully connected layers. In image classification problems, CNN is one of the best models. Therefore, we choose CNN as the classifier. At the same time, in order to make CNN perform better, we hope to make the data distribution of the training set (noisy data) as close as possible to the data distribution of the test set (clean data). Therefore, in this method, importance reweighting is used.

### 3.2.1, Convolutional Neural Network

According to research [6], Convolutional Neural Networks perform better than normal Neural Networks when the dataset is noisy. According to research [7], convolutional layers can scan the input data and detect and identify features of the data during the training process. Therefore, in this experiment, we set three convolutional layers for the FashionMNIST dataset and set four convolutional layers for the CIFAR dataset. The pooling layer can effectively reduce the size of the image, reduce the amount of calculation, and prevent overfitting [7]. Therefore, in this experiment, we set two maxpooling layers of size (2,2) for FashionMNIST and set three maxpooling layers for the CIFAR dataset. After passing through convolutional layers and maxpooling layers, the data will be flattened and then enter the fully connected layer.

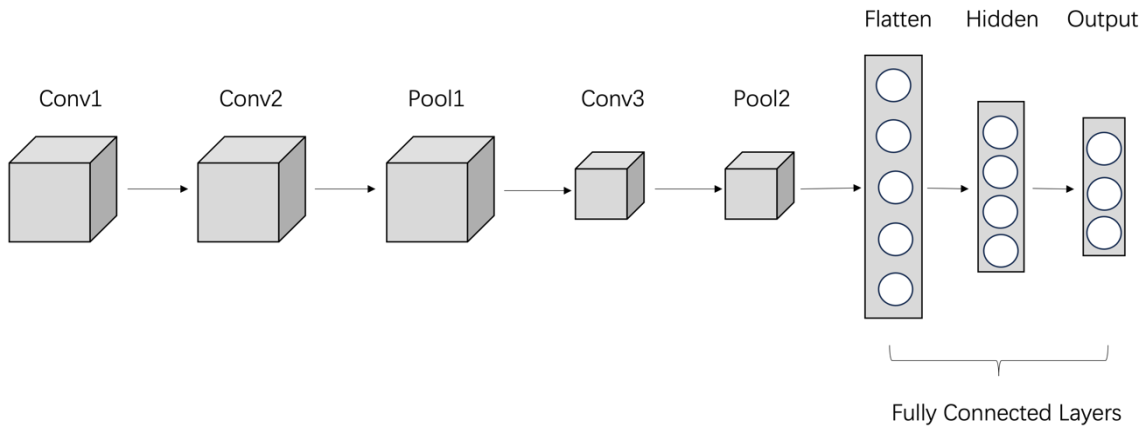

Figure 1: **Convolutional Neural Network**

In the CNN model, we use ReLU as the activation function. According to research [7], ReLU can effectively reduce the vanishing gradient problem and have a faster learning speed.

### 3.2.2, Loss Function

According to research [8], the Loss function is used to calculate the difference between the results of model prediction and the true results. Therefore, the Loss function can help the model determine the direction of optimization. In machine learning, we divide problems into classification problems and regression problems. In classification problems, we usually use Cross Entropy as the loss function of the model. Because in this experiment, the three data sets are all multi-classification problems, so according to research [9], the formula of the Loss function is:

$$Loss = -\sum_{i=1}^{M} y_i \times \log \hat{y}_i \tag{7}$$

### 3.2.3, Optimization Method

We use Adam as the optimization method. The advantage of Adam is that it can adaptively adjust the learning rate and the training speed is fast. Adam has been proven to be a very effective optimization method in neural network training. It adjusts the learning rate of each parameter by calculating the mean of the gradient ($m_t$) and the exponential moving averages of the gradient ($v_t$) [10]. So we have:

$$m_t = \beta_1 \times m_{t-1} + (1 - \beta_1) \times g_t \tag{8}$$

$$v_t = \beta_2 \times v_{t-1} + (1 - \beta_2) \times g_t^2 \tag{9}$$

$$m'_t = \frac{m_t}{(1 - \beta_1^t)} \tag{10}$$

$$v'_t = \frac{v_t}{(1 - \beta_2^t)} \tag{11}$$

The rules of update parameter:

$$\theta'_t = \theta_{t-1} = \frac{\alpha \cdot m'_t}{\sqrt{v'_t} + \epsilon} \tag{12}$$

Where θ is model parameter, β1 and β2 are hyperparameter, α is learning rate. According to research [10], β1 = 0.9, β2 = 0.999, α = 0.001.

### 3.2.4, Importance Reweighting

According to research [3], the surrogate loss function can handle data with label noise after using Importance Reweighting. And using Importance Reweighting can significantly improve the classification accuracy [1]. Because there is label noise in the training data set, importance reweighting is used to make the model perform better. There is a difference between the distribution of noisy data and the distribution of clean data, so in order to apply importance reweighting to classification problems, the expected risk of noisy data needs to approximate the expected risk of clean data.

According to research [3], we set (X, Y) as clean dataset, (X, $\hat{Y}$) as noisy dataset. Then we have:

$$P\left(\hat{Y} = +1 \middle| Y = -1\right) = \rho_{-1}, P\left(\hat{Y} = -1 \middle| Y = +1\right) = \rho_{+1} \tag{13}$$

So the expected risk of learning from clean data over noisy data:

$$R_{D,L}(f) = \mathbb{E}_{(X,Y) \sim D}[L(f(X), Y)] = \mathbb{E}_{(X,\hat{Y}) \sim D_\rho}[\beta(X, Y) L(f(X), \hat{Y})] \tag{14}$$

Where $\beta = \frac{P_D(X,Y)}{P_{D_\rho}(X,\hat{Y})} = \frac{P_{D_\rho}(\hat{Y}=y|X=x)-\rho_{-y}}{(1-\rho_{+1}-\rho_{-1})P_{D_\rho}(\hat{Y}|X)}$

Therefore, in the implementation of the code, we need to reweight the loss function according to the above principles to deal with the label noise problem in the data set. By using the Transition Matrix, we can get Clean Class Posteriro and Noisy Class Posterior.

The pseudo code is as follows:

---

**Algorithm 2** Reweight Loss Function
---
**Require:** Transition matrix $T$
   **function** REWEIGHT_LOSS($T$)
      we need calculate: $\beta(x,y) \leftarrow \frac{P_D(Y=y|X=x)}{P_{D_\rho}(Y=y|X=x)}$
      By using softmax to get probabilities of each label: $out\_softmax \leftarrow \text{softmax}(y\_pred)$

      Calculate the probabilities of true label:
      $P_D \leftarrow \text{sum}(out\_softmax \times \text{convert\_to\_float32}(y\_true\_one\_hot))$

      By using Transition Matrix to get probabilities of noisy label:
      $out\_T \leftarrow \text{dot}(out\_softmax, \text{transpose}(T))$
      $P_{D_p} \leftarrow \text{sum}(out\_T \times \text{convert\_to\_float32}(y\_true\_one\_hot))$

      Calculate $\beta$: $\beta \leftarrow \frac{P_D}{(P_{D_p}+\text{epsilon}())}$

      Calculate the weighted loss function:
      $new\_loss \leftarrow \beta \times \text{categorical\_crossentropy}(y\_true\_one\_hot, y\_pred)$
      **return** mean($new\_loss$)
   **end function**

---

### 3.2.5, Discussion

Therefore, when facing the Label noise problem, applying importance reweighting to the CNN model can effectively improve the performance of the model. By adjusting the weight of each sample during the training process, the model can learn image features better and improve image classification performance despite label noise. So Normalization + CNN model + Importance Reweighting can be a robustness classifier.

### 3.3, Method2: CNN with Backward correction

The noise robustness classifier is based on a convolutional neural network (CNN) and backward correction, which enhance its performance on the clean label data. For the data preprocessing, normalization is utilized since it is proved to be effective for image pattern recognition [14], which is likely to mitigate the effects of label noise.

### 3.3.1, Convolutional Neural Network

We choose CNN as the architecture because it has several advantages when deals with noisy labels, especially working with other techniques such as importance reweighting and backward learning [15]. In our noisy label classifier, we use three convolutional layers to identify important features and a pooling layer to reduce dimensions, helping the model to focus on patterns instead of the noise. We also build a dropout layer to prevent overfitting. Then the output will be flattened before entering the fully connected layer for classification.

### 3.3.2, Loss function

Loss function measures how good our neural network model is for our tasks, so it is crucial to choose appropriate loss functions. For the model without backward correction, we decide to use cross-entropy loss as our loss function because it is effective when adjusting model weights during the training process. For the model with backward learning [16], we choose to use a negative log likelihood (NLL) loss function. Given backward learning, the loss is corrected by multiplying inverse transition matrix, which represents the probability of the true label given the noisy label. Briefly, the backward corrected loss is the weighted sum of the losses for all possible true labels.

$$CrossEntropyLoss(y, \hat{y}) = -\sum y_i log(\hat{y}_i) \quad (15)$$

$$NLLLoss(y, log(\hat{y}_i)) = -\sum y_i log(\hat{y}_i) \quad (16)$$

Backward corrected loss:

$$l \leftarrow \left(\hat{p}(y|x)\right) = T^{-1} l\left(\hat{p}(y|x)\right) \quad (17)$$

$l$ represents the original loss function. $p(y|x)$ is the predicted probabilities of the model for each class. $T^{-1}$ is the inverse transition matrix.

### 3.3.3, Optimization Method

For optimization methods, the Adam algorithm is utilized because it is able to adjust learning rate and improve convergence. To mitigate to effect of noisy labels, the model incorporates a backward learning step during the training process in order to obtain possible true probabilities by reversing the effect of noises.

### 3.3.4, Discussion

In theory [16], our noise robustness classifier is exceptionally effective for the class-dependent noise because we go one step back in the Markov chain, described by the transition matrix T. Based on the definition and mathematical expression of backward correction, the loss correction is unbiased. Therefore, the minimizers are the same, implying that corrected loss is equal to the loss obtained on the clean data under conditional label noise. CNN + backward correction + normalization classifier is expected to be robust to noisy label.

## 4, Experiments
### 4.1, Data Analysis

In this experiment, we have three datasets for model training and testing: FashionMINIST0.5, FashionMINIST0.6, and CIFAR. We can divide datasets into three parts: training dataset, validation dataset, and testing dataset. The original dataset has already included the training dataset and testing dataset. But training dataset and the validation dataset are not split in the original data set. So we split 80% of the data set as the training data set and the remaining 20% as the validation data set. This process is completely random. When we test the performance of the model, we choose to train ten times cross-validation on the model to get more accurate results. The training set and validation set used in each training are randomly sampled. There are 3 labels in these datasets: 0, 1, and 2. On the training set and validation set, the labels of the data are noisy. On the test set, the labels of the data are clean. We can use these datasets to verify whether our model is robust under label noise.

### 4.1.1, FashionMINIST0.5

There are 21,000 samples in this datasets, and the shape of each image is (28, 28). Because the images in this dataset are graycale images, the channel is 1. The number of training dataset is 14,400, the number of validation dataset is 3,600, and the number of testing dataset is 3,000. The distribution of three labels is equal in this dataset. That means that the number of each label is the same.

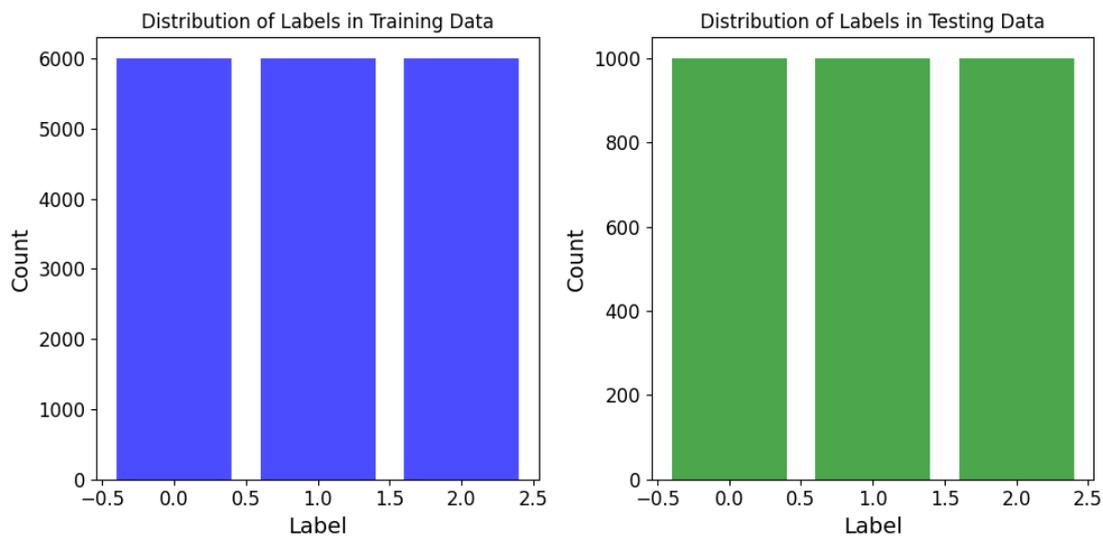

Figure 2: **Distribution of each label in FashionMINIT0.5**

According to the testing dataset, we can see the correct label. The label 0 represents T-shirt, the label 1 represents pants, and the label 2 represents skirt:

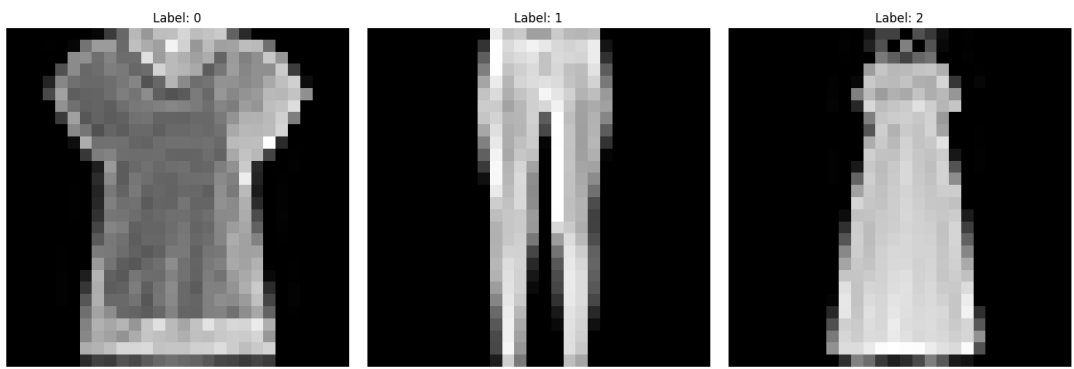

Figure 3: **Correct image of each label in FashionMINIST0.5**

When we randomly display the images in the training dataset, we will find that there is an issue where the label does not match the correct image. This is label noise:

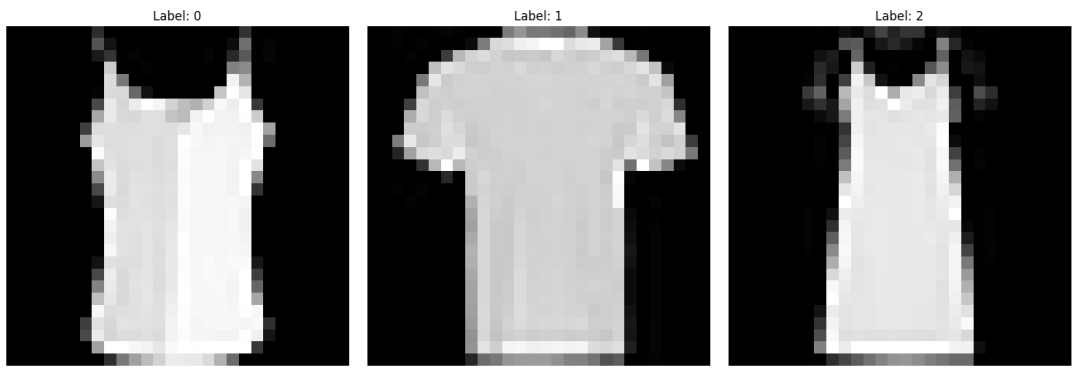

Figure 4: **Image of noisy label in FashionMINIST0.5**

**4.1.2, FashionMINIST0.6**

In this dataset, the number of training dataset, validation dataset and testing dataset is same as FashionMINIST0.5, that is the number of training dataset is 14400, the number of validation dataset is 3600, and the number of testing dataset is 3000. The shape of image is (28,28). The distribution of three label is the same.

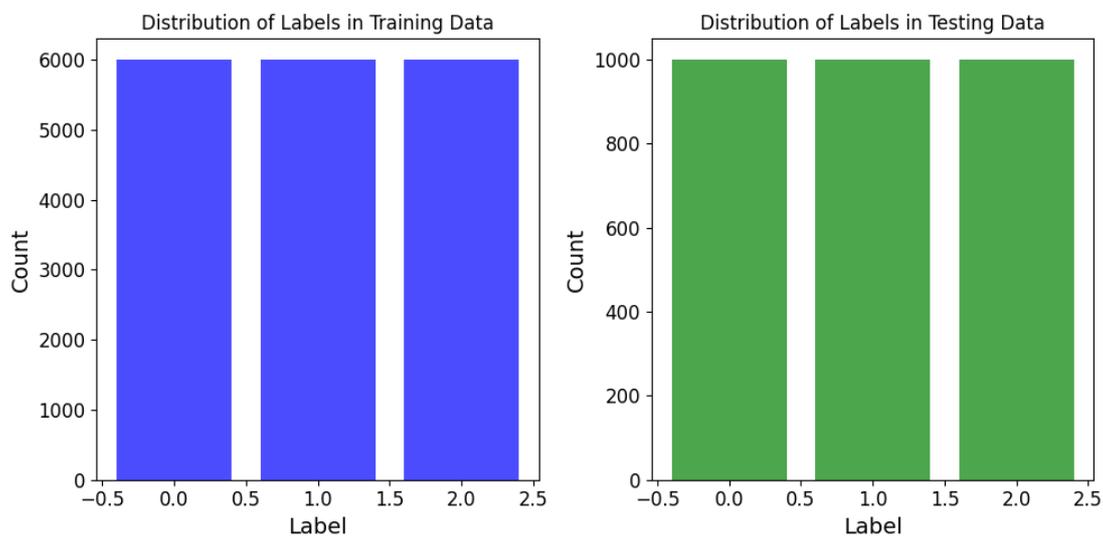

Figure 5: **Distribution of each label in FashionMINIT0.6**

Firstly, we look at the image from testing dataset (clean label):

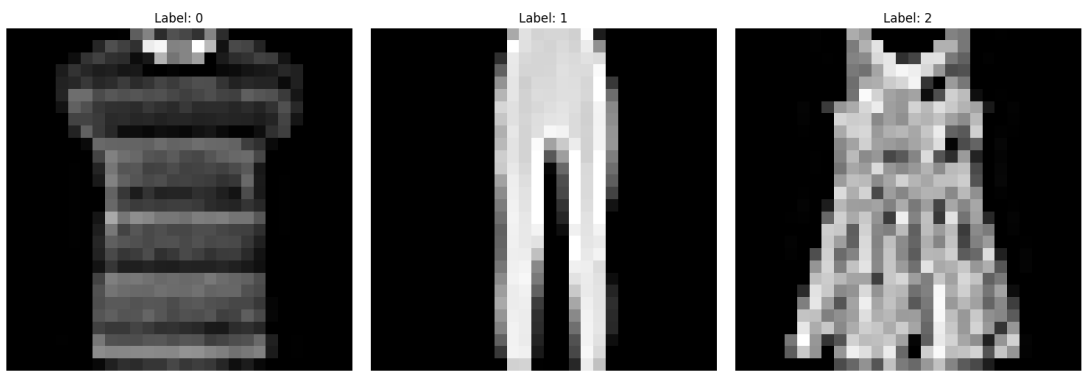

Figure 6: **Correct image of each label in FashionMINIST0.6**

Then, we can look at the image from training dataset (noisy label):

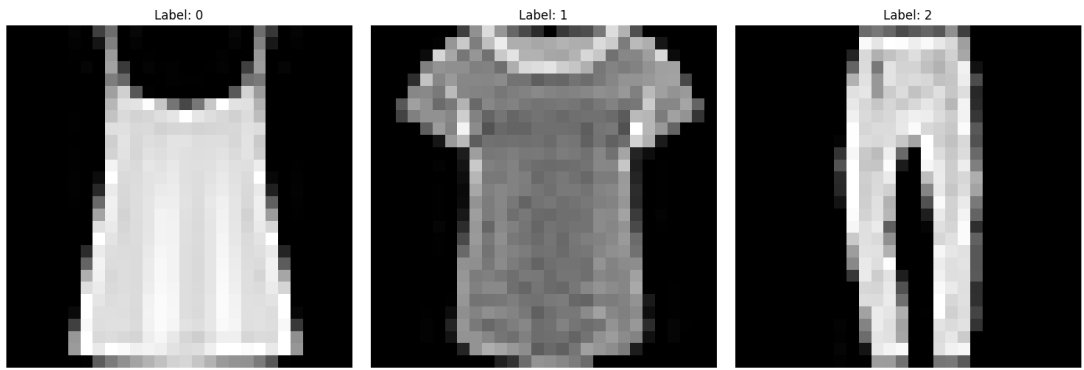

Figure 7: **Image of noisy label in FashionMINIST0.6**

The main difference between FashionMINIST0.6 and FashionMINIST0.5 is the flip rate. This part will be introduced in Estimation Method of the transition matrix.

**4.1.3, CIFAR**

There are 18,000 samples in this dataset, and the shape of image is (32, 32, 3). The images are RGB images, so the channel is 3. The number of training dataset is 12,000, the number of validation dataset is 3,000, and the number of testing dataset is 3,000. The distribution of three label is equal.

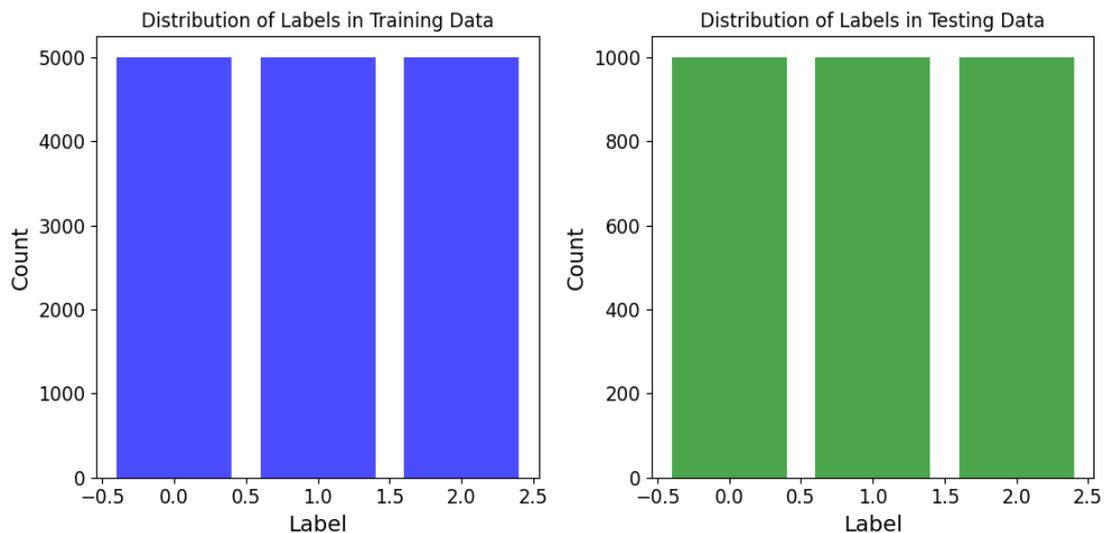

Figure 8: **Distribution of each label in CIFAR**

By looking at the clean data in the test set, we can know that when the label is 0, the image is an airplane; when the label is 1, the image is a car; when the label is 2, the image is a cat.

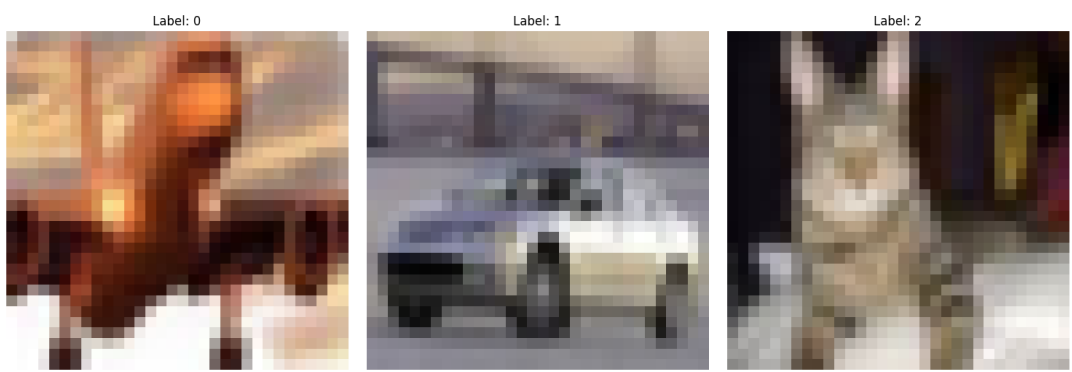

Figure 9: **Correct image of each label in CIFAR**

We can see that the dataset has misclassifications by looking at the noisy labels in the training set.

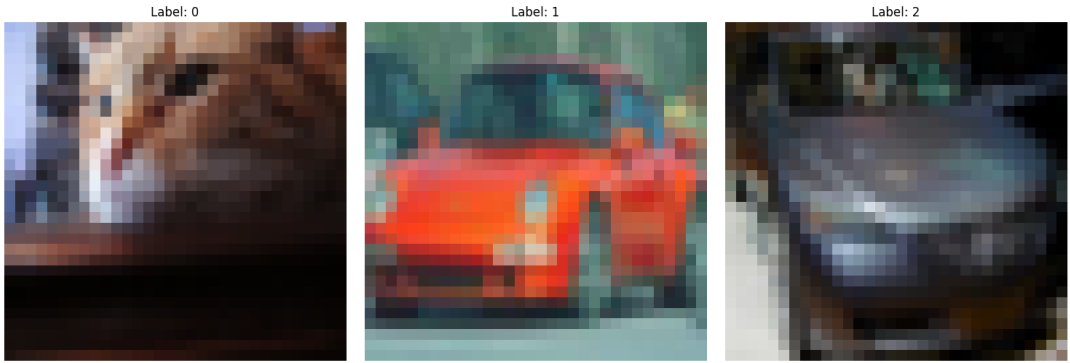

Figure 10: **Image of noisy label in CIFAR**

## 4.2 Evaluation Metrics

In order to better compare the performance of each model, in addition to calculating the Accuracy of the model, we will also use top 1 accuarcy, Precision score, Recall score, and F1 score. Firstly we will introduce the confusion matrix:

|  |  | Actual Values | |
| --- | --- | --- | --- |
|  |  | Positive | Negative |
| Predicted values | Positive | True Positive | False Positive |
|  | Negative | False Negative | True Negative |

Figure 11: **Confusion Matrix**

From the figure, we can know that TP means when the predicted value is positive and the actual value is also positive; TN means when the predicted value is negative and the actual value is also negative; FN means when the predicted value is negative and the actual value is also positive; FP means when predicted value is positive and the actual value is also negative.

- Top 1 Accuracy: The proportion of correct predictions made by the model on the entire test set. Therefore:

$$Top1Accuracy = \frac{TruePositives + TrueNegatives}{SizeOfTestSamples} \quad (18)$$

- Precision Score: It is used to evaluate the proportion of samples predicted as positive by the model that are actually positive. Therefore:

$$Precision = \frac{TruePositives}{TruePositive + FalsePositive} \quad (19)$$

- Recall Score: It is used to evaluate what proportion of all true positive classes (actual value is positive) the model predicts as positive class. Therefore:

$$Recall = \frac{TruePositives}{TruePositives + FalseNegatives} \quad (20)$$

- F1 Score: It is an indicator that comprehensively considers Recall and Precision. F1 score is very useful when dealing with imbalanced labels in the data set. Therefore:

$$F1Score = 2 \times \frac{Precision \times Recall}{Precision + Recall} \quad (21)$$

In the experiment, we will evaluate the performance of the model through these four scores. In order to more accurately evaluate the performance of the model, we will train the model 10 times and calculate the mean and standard deviation of the scores.

4.3 Evaluation transition matrix Estimator

We can estimate the transition matrix by using the method provided in 3.1 section. To evaluate the difference between predicted transition matrix and the true transition matrix, we use Mean Squared Error to calculate the difference between the predicted value and the true value.

$$MSE = \frac{1}{n}\sum_{i=1}^{n}(y_i - \hat{y_i})^2 \quad (22)$$

Because FashionMNIST0.5 and FashionMNIST0.6 provide true transition matrix. So we will first make predictions on these two data sets and evaluate the differences.

The true transition matrix of FashionMNIST0.5 is:

$$\begin{matrix} 0.5 & 0.2 & 0.3 \\ 0.3 & 0.5 & 0.2 \\ 0.2 & 0.3 & 0.5 \end{matrix}$$

The estimate transition matrix of FashionMINIST 0.5 is:

$$\begin{matrix} 0.50795323 & 0.20026277 & 0.3369517 \\ 0.29097453 & 0.51545948 & 0.24141385 \\ 0.20107204 & 0.28427809 & 0.42163846 \end{matrix}$$

The MSE is: 0.001094796813976767

The true transition matrix of FashionMNIST0.6 is:

$$\begin{matrix} 0.4 & 0.3 & 0.3 \\ 0.3 & 0.4 & 0.3 \\ 0.3 & 0.3 & 0.4 \end{matrix}$$

The estimate transition matrix of FashionMINIST 0.6 is:

$$\begin{matrix} 0.36052278 & 0.29172212 & 0.30938146 \\ 0.30907449 & 0.38835666 & 0.29762521 \\ 0.33040264 & 0.31992134 & 0.39299306 \end{matrix}$$

The MSE is : 0.00036764631834922286

The MSE is a very small value on both data sets, which shows that the differences between the predicted transition matrix and the true transition matrix are very small. Therefore, we can use it to predict the transition matrix of the CIFAR dataset.

The predicted transition matrix of CIFAR dataset:

$$\begin{matrix} 0.33922571 & 0.32896438 & 0.32101667 \\ 0.34665173 & 0.32973251 & 0.31600001 \\ 0.31412277 & 0.34130427 & 0.36298594 \end{matrix}$$

After obtaining the Transition Matrix of the CIFAR data set, we can use the same method as training the model on the FashionMINIST dataset to train the model on the CIFAR dataset. Next, we will introduce the performance of two robust classifiers we used in this experiment.

### 4.4, Method1: Normalization + CNN + Importance Reweighting

#### 4.4.1, Setup

We want to build a classifier that is robust to label noise. According to the content of Methods section, we used Preprocessing, CNN model and importance reweighitng to build this classifier.

**Preprocessing**

In data preprocessing, we used two methods: normalization and splitting the training set and validation set. Because the three data sets are all image data, we can directly divide the value of X by 255 to make the data range from 0 to 1. In the section on Datset Analysis, we explained that the data set includes training data, validation data, and test data. Among them, we split 20% of the original training set as the verification set, which can effectively avoid bias. On CNN model, we use the EarlyStopping method to avoid overfitting of the model.

**Comparison**

In order to compare whether the performance of the classifier has been improved, the experiment will be divided into 4 parts for comparison:

1. Without normalizing the data set, directly use the CNN model for training.
2. Without normalizing the data set, use the CNN model and importance reweighting methods for training.
3. Normalize the data set and use the CNN model for training
4. Normalize the data set and use the CNN model and importance reweighting methods for training.

We can know the impact of each method on the robustness of the model by using the control variable method. We will do hyperparameter tuning of the CNN model when we select the best method. By adjusting the size of filters in the convolutional layers and batch size, we can find the best combination of parameters. Then we get a robustness classifier.

#### 4.4.2, Result

The below table shows the performance scores of five classifiers. We set the Normal CNN model as the basic model, and compare it with CNN+importance reweighting, Normalization+CNN, and Normalization+CNN+importance reweighting. The Growth Rate in the table will reflect the im-provement of these classifiers compared to the basic model.

| Score | CNN | CNN + Reweighting | Growth Rate | CNN + Normalization | Growth Rate | CNN+ Reweighting + Normalization | Growth Rate | Hyper-parameter tuning (filters [32,64], BatchSize 64) | Growth Rate |
|---|---|---|---|---|---|---|---|---|---|
| Accuracy | 0.835+-0.031 | 0.892+-0.011 | 6.82 | 0.937+-0.008 | 12.18 | 0.939+-0.007 | 12.45 | 0.942+-0.004 | 12.71 |
| Top1 Acc | 0.890+-0.033 | 0.928+-0.022 | 4.26 | 0.958+-0.02 | 7.60 | 0.959+-0.016 | 7.77 | 0.961+-0.015 | 7.94 |
| Precision | 0.840+-0.063 | 0.895+-0.052 | 6.54 | 0.938+-0.037 | 11.61 | 0.941+-0.041 | 11.94 | 0.942+-0.035 | 12.12 |
| Recall | 0.835+-0.079 | 0.892+-0.046 | 6.81 | 0.937+-0.023 | 12.16 | 0.939+-0.019 | 12.44 | 0.941+-0.018 | 12.70 |
| F1 Score | 0.834+-0.051 | 0.892+-0.033 | 6.91 | 0.937+-0.023 | 12.27 | 0.939+-0.022 | 12.52 | 0.941+-0.021 | 12.81 |

Table 1: FashionMINIST0.5

| Score | CNN | CNN + Reweighting | Growth Rate | CNN + Normalization | Growth Rate | CNN+ Reweighting + Normalization | Growth Rate | Hyper-parameter tuning (filters [32,64], BatchSize 64) | Growth Rate |
|---|---|---|---|---|---|---|---|---|---|
| Accuracy | 0.706+-0.082 | 0.766+-0.077 | 8.48 | 0.884+-0.017 | 25.23 | 0.896+-0.010 | 26.93 | 0.892+-0.012 | 26.36 |
| Top1 Acc | 0.804+-0.066 | 0.844+-0.059 | 4.96 | 0.923+-0.03 | 14.76 | 0.931+-0.027 | 15.76 | 0.928+-0.028 | 15.44 |
| Precision | 0.719+-0109 | 0.771+-0.093 | 7.20 | 0.889+-0.062 | 23.58 | 0.903+-0.069 | 25.49 | 0.898+-0.067 | 24.84 |
| Recall | 0.706+-0.155 | 0.766+-0.112 | 8.48 | 0.884+-0.065 | 25.23 | 0.896+-0.054 | 26.93 | 0.892+-0.056 | 26.36 |
| F1 Score | 0.701+-0.113 | 0.765+-0.090 | 9.12 | 0.884+-0.046 | 26.18 | 0.897+-0.039 | 27.99 | 0.893+-0.041 | 27.41 |

Table 2: FashionMNIST0.6

Because the shape of the image in the CIFAR dataset is larger than the shape of the image in the two FashionMINIST datasets. According to the Transition Matrix of the CIFAR data set, it has more noise labels. Therefore, we found that if we continue to use the previous CNN model (3 convolutional layers and 2 pooling layers), underfitting will occur. The model cannot learn the data features well, resulting in poor model performance. Therefore, we need to modify the CNN architecture. We add a convolutional layer and a maxpooling layer into model. We increase the filter size of each convolutional layers from [16, 32, 64] to [32, 64, 128, 128]. Also, the number of neurons in the hidden layer is increased from 64 to 200. In the following content, this CNN model will be called Enhanced CNN.

After making these modifications, the model's performance improved significantly. The below table shows the performance of different models. The first two columns show the performance of the CNN with only 3 convolution and 2 pooling layers. The CNN model used in the remaining columns is the modified model (with 4 convolutional layers and 3 pooling layers)

| Score | CNN | CNN+ Reweighting | Enhanced CNN | Enhanced CNN+ Reweighting | Growth Rate | Enhanced CNN + Normalization | Growth Rate | Enhanced CNN + Normalization + Reweighting | Growth Rate |
|---|---|---|---|---|---|---|---|---|---|
| Accuracy | 0.4296 +- 0.065 | 0.4308 +-0.052 | 0.491 +- 0.029 | 0.536 +- 0.041 | 9.16 | 0.6043 +- 0.037 | 23.08 | 0.6329 +- 0.043 | 28.9 |
| Top1 Acc | 0.6198 +- 0.091 | 0.6205 +- 0.084 | 0.6607 +- 0.042 | 0.6907 +- 0.047 | 4.54 | 0.7362 +- 0.041 | 11.43 | 0.7553 +- 0.037 | 14.32 |
| Precision | nan | nan | 0.5029 +- 0.063 | 0.5475 +- 0.064 | 8.87 | 0.6209 +- 0.084 | 23.46 | 0.6487 +- 0.083 | 28.99 |
| Recall | 0.4286 +- 0.256 | 0.4308 +-0.256 | 0.4901 +- 0.182 | 0.536 +- 0.127 | 9.37 | 0.6043 +- 0.147 | 23.3 | 0.6329 +- 0.1345 | 29.14 |
| F1 Score | nan | nan | 0.4737 +- 0.105 | 0.5307 +- 0.070 | 12.03 | 0.5971 +- 0.079 | 26.05 | 0.6272 +- 0.071 | 32.4 |

Table 3: CIFAR

Because simple CNN models will suffer from underfitting, Precision and F1 scores cannot be calculated. Therefore, we will use the Enhanced CNN model as the basic model. Compare the performance of the remaining classifiers to the performance of the basic model.

**4.4.3, Discussion**

Both FashionMNIST datasets provide Transition Matrics. For the CIFAR dataset, we can also estimate its Transition Matrix by using the method in the 3.1 section. When we know the Transition

Matrix, we can directly use Importance Reweighting to adjust the loss function. Through the evaluation scores provided by the three tables above, we can get the following conclusions:

1. The performance of the CNN model on the two FashionMNIST datasets is good. Finding the best parameter combination through Hyperparameter tuning can make CNN a robust classifier.

2. While the performance of the Simple CNN in the CIFAR dataset is not good, the performance of the CNN improved significantly after we increased the number of layers, the size of the filter, and the number of neurons. If we had more time to tune the CNN model, we may get better results.

3. Normalization plays an important role in improving the robustness of the model.

4. With the correct Transition Matrix, using Importance Reweighting has a significant effect on improving the robustness of the model.

These results are consistent with the hypothesis we mentioned in the 3.2 section. This means that we can quickly build a robust classifier simply by using Normalization, CNN models, and importance reweighting. In this method, the Transition Matrix is necessary. Without knowing the Transition Matrix, we cannot use importance reweighting to build the loss function. However, accurately evaluating the Transition Matrix is a big challenge. For common methods, such as Anchor Point and Clusterability, there are many limitations. According to research [13], when we cannot calculate the Anchor Point, the estimation of the transition matrix will become very poor. Our Transition Matrix evaluation method does not require the use of Anchor Point. By simply building a classifier and training it using a noisy labeled data set. Then use the test set containing clean labels for evaluation to get an accurate Transition Matrix. Compared to most previous studies, our method is less restrictive and the model construction is simpler. This is an easy approach for machine learning novices to understand and use.

However, this method also has certain limitations. Firstly, we have no way to try more parameter combinations to tune the CNN model because of time constraints. This will result in us not getting the theoretically best classifier. Secondly, our evaluation method for the Transition Matrix requires a test set containing clean labels. This means that this method cannot be used if our dataset does not contain a cleanly labeled test set. Therefore, in order to make our method still work in more situations, we also need to develop a method that can predict the Transition Matrix without a clean label test set.

### 4.5, Method2: Normalization + CNN + Backward

#### 4.5.1, Setup

We build a classifier that consists of normalization, CNN model, and backward correction to be robust against label noise.

#### 4.5.2, Preprocessing

For the data preprocessing, we use two methods: normalization and train-test split. To deal with image data, it is very common to normalize the pixel to a range of 0 and 1 because it enhances image pattern recognition [14]. Although the datasets are originally consisted of trainsets and test sets, we split 20% of the trainset as the validation set to find and optimize the best model.

#### 4.5.3, Comparison

Since our goal is to create a noise robustness classifier, we will compare its performance without backward correction with its performance with backward correction. They are both evaluated by 4 metrics: precision, recall, accuracy, and F1 score. We also run it 10 times and obtain the average evaluation metrics in order to improve model's reliability and robustness.

### 4.5.4, Result

The three tables below contain results of three different datasets. Each table has average precision, average recall, average accuracy, and average F1 score, as well as their standard deviation for model with and without backward correction respectively.

|  | Average precision and std | Average recall and std | Average F1 score and $_{std}$ | Average accuracy and std |
|---|---|---|---|---|
| Normalization + CNN | 0.84927 (STD: 0.045036) | 0.84593 (STD: 0.04677) | 0.84398 (STD: 0.0483) | 0.845933 (STD: 0.04677) |
| Normalization + CNN + Backward | 0.8975 (STD: 0.0151) | 0.8912 (STD: 0.0194) | 0.8902 (STD: 0.0210) | 0.8912 (STD: 0.0194) |

Table 4: FashionMNIST0.5(Method2)

|  | Average precision and std | Average recall and std | Average F1 score and $_{std}$ | Average accuracy and std |
|---|---|---|---|---|
| Normalization + CNN | 0.84798(STD: 0.041514) | 0.83226 (STD: 0.03962) | 0.83185 (STD: 0.04048) | 0.83226(STD: 0.039625) |
| Normalization + CNN + Backward | 0.8849 (STD: 0.0096) | 0.8755 (STD: 0.0180) | 0.8757 (STD: 0.0171) | 0.8755 (STD: 0.0180) |

Table 5: FashionMNIST0.6(Method2)

|  | Average precision and std | Average recall and std | Average F1 score and $_{std}$ | Average accuracy and std |
|---|---|---|---|---|
| Normalization + CNN | 0.5951 (STD: 0.0541) | 0.5761 (STD: 0.0457) | 0.5649 (STD: 0.0509) | 0.5761 (STD: 0.0457) |
| Normalization + CNN + Backward | 0.5863 (STD: 0.0576) | 0.5652 (STD: 0.0449) | 0.5521 (STD: 0.0480) | 0.5652 (STD: 0.0449) |

Table 6: CIFAR (Method2)

### 4.5.5, Discussion

The results demonstrate that our noisy label classifier is exceptionally robust to both fashion datasets, which are greyscale images with provided transition matrix. However, the performance for the CIFAR dataset, which contains color images without transition matrix, decreases dramatically. The classifier with backward correction is even worse than the one without backward correction since it heavily relies on a proper transition matrix. If the transition matrix is problematic, the model is likely to learn nothing at all during the training process. Regarding the fact that the condition number for our estimated transition matrix is too high, which leads to complete failure of our model, we decide to mix T with the identity matrix before inversion [16]. Even though the result implies that our backward correction model is still not ideal on this dataset, it at least learns about noisy label with the modified inverse transition matrix during the training process. Another limitation is that backward correction is not practical for complex noise since it assumes Markov chain for the label noise. We will try to improve the performance of our classifier on the CIFAR dataset and real-word data by adjusting the CNN architecture and exploring more suitable estimated transition matrix.

## 5, Conclusion and future work

Overall, our main goal is to study and improve the robustness of label noise classifiers, especially in the context of image classification. We strive to improve the robustness of label noise classifiers by employing adversarial machine learning (AML) and importance reweighting techniques, combined with convolutional neural networks (CNN) as baseline models. The experiments involve three different datasets - CIFAR, FashionMnist0.5 and FashionMnist0.6 and are evaluated using four key metrics such as accuracy, recall, precision and F1 score. We train multiple classifiers by applying these methods separately on three datasets. During the experiments, we adjusted the parameters of the algorithm and monitored the training process of the model to ensure that it could effectively handle label noise on different datasets.

Our experiments show that adversarial machine learning and importance reweighting methods indeed significantly improve the robustness of label noise classification performance. Although these methods often perform with mediocre results on CIFAR datasets, after being preprocessed and Transition Matrix. Both methods are still effective in mitigating the impact of label noise. The experimental results obtained by the two methods show that the CNN model performs well on both FashionMNIST data sets. Finding the best combination of parameters through hyperparameter tuning can make CNN a powerful classifier.

Despite the positive results of the experiment, there are still some potential experimental design flaws. Future research directions may include increasing the size of the dataset and introducing more domains and scenarios to more comprehensively evaluate the model's performance. In addition, future research should explore wider data sets and more complex noise models to comprehensively evaluate the robustness of the algorithm. In addition, parameter optimization and importance reweighting methods of adversarial machine learning can further improve performance. More advanced adversarial machine learning techniques, such as generative adversarial networks (GAN), can be considered to further improve the robustness of the model [17]. For importance reweighting methods, one can delve into more complex weighting strategies.

Finally, our paper not only addresses the problem of label noise, but also highlights the ongoing developments in the field of supervised learning. As we face the complexity of real-world data, the pursuit of better robust, adaptable models remains critical, thereby pushing the field in a more reliable and applicable direction.